\documentclass[conference, compsocconf]{IEEEtran}
\IEEEoverridecommandlockouts
\usepackage{cite}
\usepackage{amsmath,amssymb,amsfonts}
\usepackage{algorithmic}
\usepackage{graphicx}
\usepackage{textcomp}
\usepackage{xcolor}
\usepackage{arydshln}
\usepackage[caption=false]{subfig}
\def\BibTeX{{\rm B\kern-.05em{\sc i\kern-.025em b}\kern-.08em
    T\kern-.1667em\lower.7ex\hbox{E}\kern-.125emX}}

\DeclareRobustCommand*{\IEEEauthorrefmark}[1]{%
  \raisebox{0pt}[0pt][0pt]{\textsuperscript{\footnotesize #1}}%
}

\begin{document}

\title{Egocentric Hand Track and Object-based Human Action Recognition}

\author{\IEEEauthorblockN{Georgios Kapidis\textsuperscript{*}\thanks{* Corresponding author: georgios.kapidis@noldus.nl, g.kapidis@uu.nl}\IEEEauthorrefmark{1}\IEEEauthorrefmark{2}, Ronald Poppe\IEEEauthorrefmark{2}, Elsbeth van Dam\IEEEauthorrefmark{1}, Lucas P. J. J. Noldus\IEEEauthorrefmark{1}, Remco C. Veltkamp\IEEEauthorrefmark{2}}
\IEEEauthorblockA{
\IEEEauthorrefmark{1}Noldus Information Technology, Wageningen, The Netherlands\\
\IEEEauthorrefmark{2}Department of Informatics and Computer Science, University of Utrecht, Utrecht, The Netherlands}
}


\maketitle

\begin{abstract}
Egocentric vision is an emerging field of computer vision that is characterized by the acquisition of images and video from the first person perspective. In this paper we address the challenge of egocentric human action recognition by utilizing the presence and position of detected regions of interest in the scene explicitly, without further use of visual features.

Initially, we recognize that human hands are essential in the execution of actions and focus on obtaining their movements as the principal cues that define actions. We employ object detection and region tracking techniques to locate hands and capture their movements. Prior knowledge about egocentric views facilitates hand identification between left and right. With regard to detection and tracking, we contribute a pipeline that successfully operates on unseen egocentric videos to find the camera wearer's hands and associate them through time. Moreover, we emphasize on the value of scene information for action recognition. We acknowledge that the presence of objects is significant for the execution of actions by humans and in general for the description of a scene. To acquire this information, we utilize object detection for specific classes that are relevant to the actions we want to recognize. 

Our experiments are targeted on videos of kitchen activities from the Epic-Kitchens dataset. We model action recognition as a sequence learning problem of the detected spatial positions in the frames. Our results show that explicit hand and object detections with \textit{no} other visual information can be relied upon to classify hand-related human actions. Testing against methods fully dependent on visual features, signals that for actions where hand motions are conceptually important, a region-of-interest-based description of a video contains equally expressive information with comparable classification performance. 
\end{abstract}


\section{Introduction}
In recent years the egocentric point of view has been employed by the research community to address computer vision challenges such as activity recognition \cite{fathi_understanding_2011} and object detection \cite{fathi_learning_2011} traditionally contemplated as belonging in the domain of third-person vision. Since then, egocentric vision has been applied to more elaborate applications including video summarization \cite{lee_discovering_2012} and social interaction analysis \cite{yonetani_recognizing_2016}. Notably, it has also expanded into the domain of health-care \cite{doherty_wearable_2013} where static camera systems tend to struggle to a greater extent following privacy concerns \cite{townsend_privacy_2011}. Ultimately, egocentric vision is affiliated with the domain of augmented reality towards human-centered applications that provide task-specific assistance \cite{leelasawassuk_automated_2017}, thus enhancing human independence; an appropriate fit for scenarios where human ability is impaired or reduced.

The prominent characteristic of egocentric vision is that it provides a first person perspective of the scene by placing a forward-facing wearable camera on the chest or head of a human. This provides a unique view that is person-centric and optimally set to capture information that is arguably more relevant to the camera wearer \cite{kanade_first-person_2012}. Naturally, this refers to the surrounding area and its contents, usually consisting of objects, hands, other people and the scene background. Being able to examine a perspective of the scene that accumulates all this information with clarity allows for improved inference of higher level cues such as the quantification of interactions between the hands from their proximity \cite{nguyen_recognition_2018}, object-activity relations from associated movements \cite{bertasius_first_2017} and location identification from the presence of distinctive objects \cite{kapidis_where_2018}. 

\begin{figure}[b]
    \centering
    \includegraphics[width=120pt]{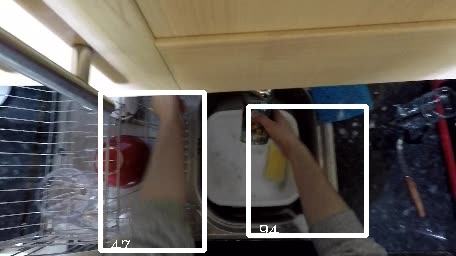}
    \includegraphics[width=120pt]{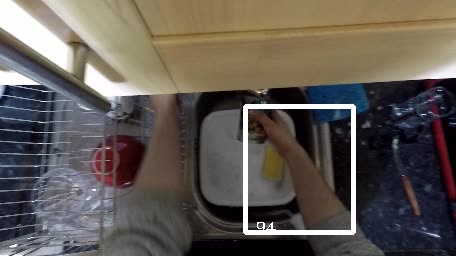}
    \caption{Results of a hand detector from two consecutive frames on the Epic-Kitchens dataset \cite{damen_scaling_2018}.}
    \label{fig:fig0}
\end{figure}

However, understanding of visual content in human intelligible terms remains challenging despite being facilitated from the egocentric perspective. That is, recognition of a specific area in view as the `hand` or as a specific object class which may or may not be under manipulation, will always deteriorate due to inter-object and object-hand occlusions. In the video domain the recognition task can become more challenging in the egocentric setup, due to motion noise introduced by rapid movements of the camera or of the objects seen from this perspective. In Figure~\ref{fig:fig0} we illustrate an example where slight motion of a hand causes missed detections. 

This limitation poses a challenge for algorithms developed in the context of third-person vision in terms of their applicability into a comparable yet divergent field. We argue that methods yielding cues towards human-like understanding of a scene from one domain can be compatible with egocentric vision given a certain amount of fine-tuning. The scope of this work is to assess up to what point existing object detection and tracking schemes can produce valuable information for egocentric action recognition. Our idea relates to methods that \textit{reduce} RGB images to trajectories or poses of hands or objects in the scene and use this contextual information alone for human action recognition \cite{baradel_human_2017, nguyen_recognition_2018} or a related task such as prediction \cite{furnari_next-active-object_2017}. Our objective is to investigate the information encoded in egocentric \textit{movements} of hands and objects in contrast to the currently predominant approach of using the visual information directly \cite{carreira_quo_2017}. Are the motion sequences alone able to be the basis for modeling actions? Our aim is to test the limits of object detection and tracking methods in their ability to produce usable data towards higher level inference.

Initially, we focus on distinguishing the action specific cues that can be acquired solely from hand movements. Instinctively, human hand movements are expected to carry much of the spirit of actions that are explicitly named after the actual motion itself e.g. 'put', 'take', 'stir', 'open', 'close' etc. We strive to exploit the clear view of the hands and their movements in egocentric videos and study them closely towards identifying associated actions, facilitated by detection and tracking of hand regions. Furthermore, we capitalize on the structure of the actions themselves which are generally not only associated with the hand movements but also related to objects of interest arising from the context of the scene. For example 'wash dishes' as in Figure~\ref{fig:fig0}.

\begin{figure}[ht]
\centerline{\includegraphics[width=250pt]{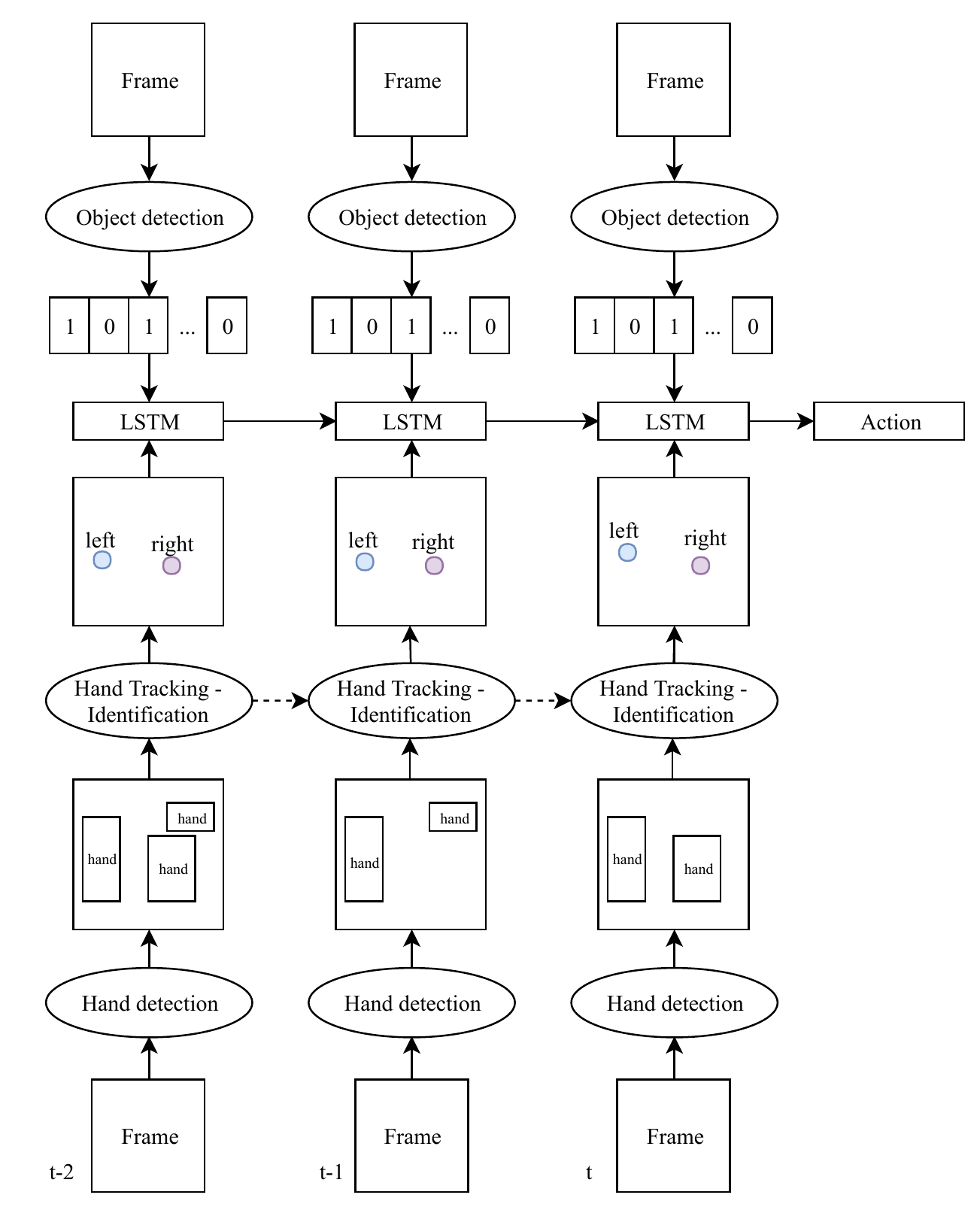}}
\caption{Our pipeline for action recognition. Hands are detected with YOLOv3 \cite{redmon_yolov3:_2018} in Epic-Kitchens \cite{damen_scaling_2018} and tracked by detection using SORT \cite{bewley_simple_2016}. We remove track overlaps and identify left and right hands from their position on the frame. For the objects we also rely on YOLOv3 with a second model trained on the noun classes of Epic-Kitchens. Binary presence vectors are propagating the object knowledge per frame. Finally, hand and object information is used as input for the LSTM to classify actions.}
\label{fig:fig1}
\end{figure}

This work is directly associated with the production of hand trajectories. The prelude is that an object detector is applied on egocentic videos with the aim to accurately detect the hands, thus substituting the arduous task of manually labelling hand regions, to an automated process. Subsequently, tracking is applied to temporally associate the detections into meaningful sequences, which are cleared from overlapping misdetections and attributed to the left or the right hand. Finally the hand trajectories are augmented with the presence of objects and used as input for action classification. Figure~\ref{fig:fig1} illustrates our approach. 

The contributions of this paper are threefold:
\begin{itemize}
    \item A hand detection, tracking and identification pipeline that extracts hand motions from egocentric videos, structured to provide the hand positions for every frame including the distinction into left and right hands.
    \item The assessment of the capability of hand tracks alone for egocentric action recognition and the effects of temporal sampling in the representative ability of hand motions.
    \item Further experimentation with the inclusion of object presence and position to capitalize on the limited set of specific actions that an object can be associated with.
\end{itemize}

In Section~\ref{sec:related} we discuss related work about action recognition in egocentric videos with a focus on hand-object interactions. In Section~\ref{sec:dataset} we describe our hand detection, tracking and identification pipeline, in Section~\ref{sec:lstm} the temporal classification problem for action recognition and in Section~\ref{sec:experiments} our experiments and results. Finally, in Sections~\ref{sec:discussion} and \ref{sec:conclusions} we discuss our findings and conclude the paper.

\section{Related Work}
\label{sec:related}
\subsection{Egocentric activity and action recognition}
One of the first works about egocentric activity understanding focuses on the intrinsic information that defines the egocentric vision paradigm \cite{fathi_understanding_2011}. They modelled relationships between hands, objects and actions using extracted visual features to model activities and showed through bottom-up and top-down models the mutual improvements that these relationships offer. We base our work on the concept that hands and objects are rudimentary for egocentric action recognition and video understanding. 

Motion based egocentric action recognition is described in \cite{li_delving_2015}. Hand-object interaction points, objects, head and ego-motion are classified into actions based on motion and color-based features and trajectories extracted from the video frames, but without a specific focus on modeling the hand locations. Another multimodal approach for egocentric activity recognition is described in \cite{ma_going_2016}. Here, a hand segmentation network, an object localization network and a network trained on motion flow are joined to predict actions. Their understanding that hands and objects are fundamental for actions is similar to ours, but our pipelines are different in that we model the temporal associations of the tracked hands and objects in the video instead of structured feature representations of the raw image pixels with CNN features or other descriptors.

In \cite{bertasius_first_2017} a two-stream visual segmentation-based architecture is used to predict the interaction areas between hands and objects in a video stream and model them as actions. The concept of hand-object interactions is further explored in \cite{cai_understanding_2016} with the detection of grasps in relation to the shape of objects for modeling actions. End-to-end methods also include \cite{baradel_object_2018}, where in order to recognize actions a network is trained on pairs of frames and is jointly optimized for the training objectives of action recognition, object segmentation and inter-frame object interactions and their temporal association with recurrent networks. We also match the intuition that hands and objects are fundamental for egocentric actions, however we emphasize on the explicit detection of hand and object regions and their positions to recognize actions.

\subsection{Hands and objects}
The explicit exploration of hands and objects and their temporal associations has seen a significant volume of work in the egocentric action recognition domain. Initially, hand detection, segmentation and identification techniques \cite{li_pixel-level_2013, betancourt_sequential_2014, betancourt_left/right_2016} are developed and their results are utilized to model actions or activities. In this work we rely on single frame object detectors for both hands and objects.

Specifically hand based activity recognition from the egocentric perspective is discussed in \cite{bambach_lending_2015}. The EgoHands dataset together with an egocentric hand detection and segmentation pipeline are developed before inference of activities. It is one of the first works that show the difference between relying on hand detection or segmentation and using manual labels for activity classification. Inspired by this, we augment the EgoHands hand dataset with additional samples for our hand detector and introduce tracking into the action recognition pipeline to improve the detection output. 

Egocentric hand or object trajectories are considered for classification in \cite{furnari_next-active-object_2017, garg_drawinair:_2019, nguyen_recognition_2018}.
In \cite{furnari_next-active-object_2017} the trajectories of detected objects are computed and then classified as active or passive based on hand manipulations towards them. The detection and tracking methods are related to our scope however we focus on the particular objective of recognizing hand-based actions with the help of objects. In \cite{garg_drawinair:_2019} the fingertip positions are used as a means to identify human gestures with bidirectional LSTMs. Sequences of fingertip coordinates are classified into a predefined set of gestures. In contrast, we are interested in the whole hand and arm regions and do not rely on a predefined set of trajectories but utilize tracking to produce them. Egocentric hand based activity recognition is considered in \cite{nguyen_recognition_2018} where the distance between detected hands or the distance between detected hands and objects marked as active are considered as features for activity classification. In this work, we capture the trajectory of each hand and detect the objects instead of using manual annotations, in order to introduce real-world complications such as unstable tracks and false detections to eventually improve robustness.

\section{Hand Track Dataset} 
\label{sec:dataset}


In this section we describe the process to produce a hand track dataset from the raw frames of Epic-Kitchens\cite{damen_scaling_2018}. Our aim is to capture the position of each of the (at most) two visible egocentric hands in view as an (x,y) coordinate for every video frame. The coordinate pair signifies the center of the bounding box of a detected hand.

\subsection{Epic-Kitchens}
The Epic-Kitchens dataset comprises a set of 432 egocentric videos recorded by 32 participants in their kitchens at 60fps with a head mounted camera. There is no guiding script for the participants who freely perform activities in kitchens related to cooking, food preparation or washing up among others. Each video is split into short action segments (mean duration is 3.7s) with specific start and end times and a verb and noun annotation describing the action (e.g. `open fridge`). The verb classes are 125 and the noun classes 352. The dataset is divided into one train and two test splits.
~For both test sets, the verb and noun annotations are not yet openly available
~hence we focus our work on the fully annotated train set (272 videos, 28 participants). We partition it into custom train and test splits based on the participant ids\footnote{Videos from participants 1-8, 10, 12-17, 19-24 are our train and 25-31 our test set respectively.}
~to avoid videos from the same person in both splits. Additionally, almost 300k object bounding boxes are provided for the videos of the original train set which we utilize to train an object detector (Section~\ref{sec:objdet}).

For the rest of the paper our subset of Epic-Kitchens is referred to as Epic-Kitchens, unless stated otherwise.


\subsection{Hand detection with YOLO}
In order to acquire hand regions from Epic-Kitchens we train a hand detector with YOLOv3\cite{redmon_yolov3:_2018} on the combination of a collection of egocentric hand datasets.
\subsubsection{Dataset collection}

We utilize hand annotations from existing egocentric datasets, namely EgoHands \cite{bambach_lending_2015}, EGTEA Gaze+ \cite{li_eye_2018, li_delving_2015}, CMU EDSH \cite{li_pixel-level_2013} and THU-READ\cite{tang_action_2017}.
Since we are interested in detection and not segmentation we only keep the bounding rectangle of a hand mask and use this as the ground truth for a hand region. Next, we augment the dataset with negative samples i.e. frames that do not contain visible hands or annotations, in order to punish the objectness learning part of the network and produce fewer false proposals, ultimately reducing false positive detections. We manually annotate 11,683 such frames from the Intel Egocentric Object Recognition Dataset \cite{ren_egocentric_2009}. Information about the amount of hand annotations and the size of each train and test split is detailed in Table~\ref{tab:hand_datasets}.

\begin{table}[htbp]
    \centering
    \caption{Collection of hand annotations from egocentric datasets}
    \begin{tabular}{l|c|c|c:c}
        Dataset & Images & Hand annotations & Train & Test \\
        \hline
        EgoHands \cite{bambach_lending_2015} & 4,800 & 14,884 & 11,440 & 3,444 \\
        Egtea Gaze+ \cite{li_eye_2018} & 13,847 & 15,258 & 14,295 & 963 \\
        CMU EDSH \cite{li_pixel-level_2013} & 743 & 1,394 & 1,186 & 208 \\
        THU-READ\cite{tang_action_2017} & 652 & 1,331 & 1,252 & 79 \\
        IEORD\cite{ren_egocentric_2009} & 11,683 & 0 & - & - \\
        \hline
        Combined & 31,725 & 32,867 & 28,173 & 4,694
    \end{tabular}
    \label{tab:hand_datasets}
\end{table}

\begin{table*}[!ht]
    \centering
    \caption{Average Precision (\%) with 25\% IoU threshold. Row-wise the dataset used for training and column-wise the dataset used for testing. In parentheses False Detection Rates (\%) for each test set. (For IEORD we report the false positive counts per detector)}
    \begin{tabular}{c||c:c|c|c|c|c}
         & IEORD \cite{ren_egocentric_2009} & EDSH \cite{li_pixel-level_2013} & EgoHands \cite{bambach_lending_2015} & EGTEA+ \cite{li_eye_2018} & THU-READ \cite{tang_action_2017} & Combined Test
         \\
         \hline
        EDSH & 71 & 100 (0) & 17 (49) & 72.5 (21) & 86.17 (9) & 31.98 (34) \\
        EgoHands & 15 & 26.09 (53) & 90.58 (2) & 43.42 (39) & 17.15 (79) & 77.29 (13) \\
        EGTEA+ & 37 & 89.91 (4) & 20.51 (51) & 89.65 (5) & 74.58 (7) & 41.05 (32) \\
        THU-READ & 191 & 90.62 (1) & 19.24 (50) & 65.85 (19) & 100 (0) & 33.07 (38)\\
        Combined Train & 18 & 100 (0) & 90.53 (3) & 89.38 (7) & 90.05 (5) & \textbf{90.42 (4)}
    \end{tabular}
    \label{tab:object_detectors}
\end{table*}

\subsubsection{Training}
\label{sec:hand_det}
We perform various experiments to train the hand detectors to determine the optimal dataset combination that supports generalization, since our eventual task is to apply the detector on an unseen dataset for extraction. We train a detector for each available hand dataset (except IEORD) and one for the combined train sets. All detectors are trained for a single target class `\textit{hand}` with batch size 64, starting learning rate 10\textsuperscript{-3}, momentum 0.9 and weight decay 5*10\textsuperscript{-4}. Weights are pretrained on Imagenet\cite{deng_imagenet:_2009} and MSCOCO\cite{lin_microsoft_2014}. We do not recalculate box anchors for our datasets after preliminary tests suggesting minor performance decline. Training takes place for multiple shapes with starting input dimensions of the detector 416x416. We evaluate all detectors on each test set using Average Precision (AP\textsubscript{25}) and False Detection Rate\footnote{Since we plan to use the detections as a preamble for tracking, the frequency of false positive detections is an important metric to consider.} (FDR = 1-Precision). In Table~\ref{tab:object_detectors} we illustrate the best performing weights of each detector, based on the two metrics. We show that in terms of both AP and FDR, the detector based on the dataset combination performs best.


\begin{figure*}
    \begin{center}
        \includegraphics[width=516pt]{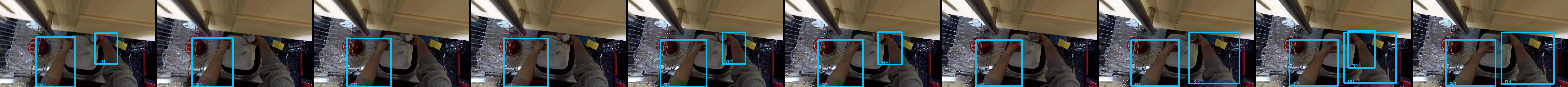}
        \vskip 1pt
        \includegraphics[width=516pt]{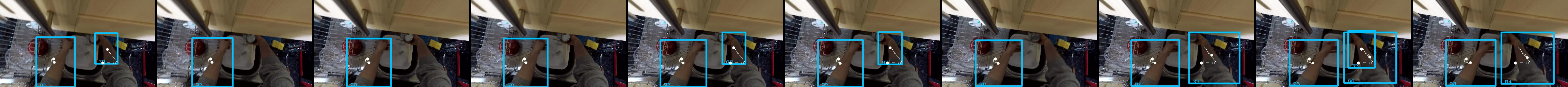}
        \vskip 1pt
        \includegraphics[width=516pt]{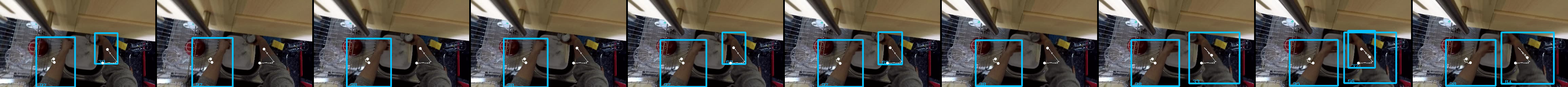}
        \vskip 1pt
        \includegraphics[width=516pt]{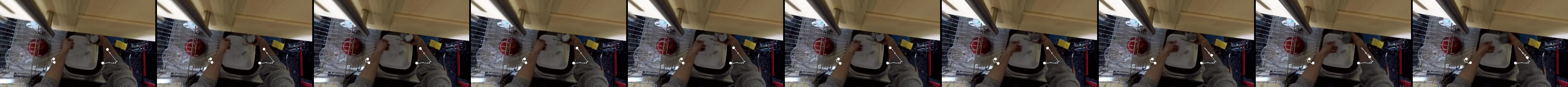}
    \end{center}
    \hskip 21pt (1) \hskip 36.5pt (2) \hskip 36.5pt (3) \hskip 36.5pt (4) \hskip 36.5pt (5) \hskip 36.5pt (6) \hskip 36.5pt (7) \hskip 36.5pt (8) \hskip 36.5pt (9) \hskip 34pt (10)
    \caption{Visualization of the results of hand tracking and track augmentation on a sequence of 10 frames. Each row showcases an additional step. Row 1 shows the detection output from Yolo. In row 2, SORT\cite{bewley_simple_2016} associates the available detections to distinct tracks. A track for each hand covers the full sequence but frames 2-4 and 7 are still not assigned with a right hand coordinate. For these frames we interpolate from the last available coordinate to the latest. In row 3, the right hand track is augmented and the missing detections are covered. In row 4 we show a case of removing a redundant right hand track on frame 9.}
    \label{fig:fig2}
\end{figure*}

\subsubsection{Detection on Epic-Kitchens}
\label{sec:handdet-epic}
We apply the combined hand detector on the Epic-Kitchens dataset to extract hand instances. We accept hand detections with confidence greater than 25\%. In Figures~\ref{fig:fig0} and \ref{fig:fig1} (row 1) we show that the detector generalizes in unseen images, however slight changes over the course of video frames that are introduced during hand movements, strong ego-motion, changing lighting conditions or occlusions can cause missed detections. Another issue we address is that of overlapping detections for the same hand regions. 

\subsection{Hand tracking with SORT}
\label{sec:tracking}
The continuity of the visual information that exists in temporal video streams enables the utilization of tracking methods to enhance the missing detections. The object detector operates on a per frame basis whereas tracking by detection combines information from multiple frames. We utilize Simple Online and Real-time Tracking (SORT)\cite{bewley_simple_2016} for this task. It associates the detections over the course of a video with threshold-based tolerance to missed ones. Hand bounding boxes are associated through the course of frames and identified as belonging to a track with a certain id. SORT uses the Kalman filter\cite{kalman_new_1960} to predict the coordinates of what would likely be the next bounding box of an existing track and the Hungarian algorithm \cite{kuhn_hungarian_1955} to assign the detections from subsequent frames to existing tracks or new ones. 
Tracking with SORT is controlled by three parameters; IoU\textsubscript{min}, T\textsubscript{LOST} and T\textsubscript{min}. IoU\textsubscript{min} is the minimum required overlap between a new detection and the predicted target for a track that leads to the detection's assignment to it. T\textsubscript{LOST} defines the number of frames that a track can survive for without an assigned detection before being finalized. T\textsubscript{min} is the minimum amount of consecutive detections required to instantiate a new track or recover it after empty frames. We set IoU\textsubscript{min} to 10\% to promote track continuity against strong ego or hand motions, T\textsubscript{LOST} to 10 frames without assignment to allow for sufficient time (equals 167 ms in the 60fps videos of Epic-Kitchens) to re-establish a track and T\textsubscript{min} to 1 to revive a track instantly. In row 2 of Figure~\ref{fig:fig2} we show the instantiated tracks as points in the centers of bounding boxes.

\subsubsection{Track interpolation}
\label{sec:trc_interp}
In order to extend our knowledge about the location of the hands we introduce the concept of \textit{intermediate frames}. We define them as the video frames that are implicitly included in a track by means of previous and future frames that contain a detection for it. Intermediate frames do not hold a detection for the track, however given the inherent continuity of information in sequential video frames and the short time span we allow for a track to be kept alive without a detection, we assume that the hands do exist in these frames but are missed from the object detector. To augment the tracks for these frames we apply linear interpolation on the tracked box centers from the (x,y) coordinate of the last frame to that of the latest one. In row 3 of Figure~\ref{fig:fig2} we display frames 2-4 assigned with coordinates for the right hand.

\subsubsection{Track elimination}
In the videos of Epic-Kitchens the participants undertake kitchen activities alone. This limits the maximum number of co-occurring hand tracks at any given moment to two, one for each hand. Particularly, we assign each track to the left or the right hand of the participant based on the location of the center of the first detection of a track. Overlapping tracks for the same image region that have been associated with the same hand are removed and the longest track survives, as in Figure~\ref{fig:fig2} frame 9 with the elimination of the second track for the right hand. Finally, for the frames with no available detection and track information we assume a hand position below the view of the camera.

\subsection{Noun Object Detector}
\label{sec:objdet}
To study the hand-object relationships we require information about object presence. Epic-Kitchens includes sparse object labels for the majority of its noun classes. We utilize them to train an object detector using YOLOv3 with the same parameters as those in Section~\ref{sec:hand_det} except for the base network dimensions which are increased to 608x608 and the introduction of Sparse Pyramid Pooling \cite{he_spatial_2015} to the model structure to enhance the detector's ability to find smaller objects. We train for 50k iterations with average loss stabilized around 0.64. We apply the detector on Epic-Kitchens and accept detections with confidence greater than 25\%. 
\section{Motions to Actions}
\label{sec:lstm}
We aim to develop a frame-wise correspondence between every image of Epic-Kitchens and the hand detection tracks in order to exchange the visual information with our representation. The pipeline of Section~\ref{sec:dataset} contributes knowledge about the hand locations regardless if they are detected in a given image or not. This continuous evolution of positions leads to a sequence of coordinates that in the temporal dimension capture the motions of the hands. To gain knowledge about these motions we formulate the problem of hand track classification as a \textit{sequence learning problem}. Long Short-term Memory (LSTM) networks \cite{hochreiter_long_1997} have shown ability to model long-term dependencies in sequences of arbitrary sizes of coordinate \cite{nguyen_recognition_2018, garg_drawinair:_2019} or object presence \cite{kapidis_where_2018} data and we employ them for their classification into actions.

\section{Experiments and Results}
\label{sec:experiments}
We construct a series of experiments to demonstrate the ability of the hand track and object presence information to substitute the visual information of the raw RGB frames, in order to model actions related to hands in the egocentric perspective.

In our experiments we apply LSTM with a Fully Connected layer on the last hidden state from the final LSTM layer to obtain a class prediction for an action segment. To train the models we use cyclical learning rate (CLR) \cite{smith_cyclical_2017, smith_disciplined_2018} with a triangular policy that fluctuates between the base and the maximum learning rate in 20 epochs. Batch size is set to 128. We train our models for 1,000 epochs capitalizing on the ability of the triangular CLR policy to move weights out of local minima in search of better configurations. We use categorical cross entropy to calculate the loss and Stochastic Gradient Descent for optimization. Our learning scheme targets the 125 verb classes of the Epic-Kitchens dataset. For the LSTM experiments we consider the following features:
\begin{itemize}
    \item The \textbf{concatenated Left/Right hand coordinates} (LR) (x,y) of the center of each hand normalized to the image size. The values are in the [0-1] range when there is a hand present. Alternatively, the (x,y) coordinate is set to (0.25, 1.5) to declare that the left hand is out of view and correspondingly to (0.75, 1.5) for the right hand. Feature length is 4.
    \item The \textbf{Binary Presence Vector} (BPV) of objects \cite{kapidis_where_2018} consisting of zeros and ones with length equal to the number of the noun classes of Epic-Kitchens (352). The BPVs are concatenated to the hand coordinates for every frame and the feature length increases to 356 (352 + 4).
    \item The \textbf{tracked object coordinates} (Obj) instead of the BPV of the objects in a video frame. This increases the feature length to 708 (352*2 + 4). In case of multiple instances of an object in a frame we only consider its longest running occurrence following the tracking scheme of Section~\ref{sec:tracking}. When an object is not present on a frame its coordinates are set to (0,0).
\end{itemize}

We report classification results for our test set in Table~\ref{tab:results} in terms of overall Top1 and Top5 accuracy. Following the evaluation scheme of Epic-Kitchens we additionally report mean per-class precision and recall for verb classes with more than 100 samples during training, which in our splits are 24.

\begin{table*}[!ht]
    \centering
    \caption{Verb classification results on Epic-Kitchens}
    \begin{tabular}{l|l|l|c|c|c|c||c|c|c|c|c}
        & \multicolumn{6}{c||}{Model Parameters} & \multicolumn{2}{c|}{Accuracy \%} & \multicolumn{2}{c|}{Average \%} & \\
        \# & Model & Feature & Hidden & Layers & Seq. Length & Target & Top-1 & Top-5 & Cls Precision & Cls Recall & Epoch \\
        \hline
        1 & LSTM & LR$^{\mathrm{a}}$ & 32 & 2 & Full & Verbs & 31.100 & 74.115 & 11.02 & 10.46 & 918\\
        2 & LSTM & LR & 16 & 2 & 32 & Verbs & 31.013 & 73.148 & 10.38 & 8.46 & 628\\
        3 & LSTM & LR+BPV$^{\mathrm{b}}$ & 32 & 2 & Full & Verbs & \textbf{34.968} & \textbf{76.084} & 15.08 & 12 & 440 \\
        4 & LSTM & LR+BPV & 16 & 2 & 32 & Verbs & 34.053 & 75.358 & 17.64 & 10.83 & 553\\
        5 & LSTM & LR+Trc BPV & 16 & 2 & 32 & Verbs & 34.053 & 75.289 & 19.08 & 11.51 & 620\\
        6 & LSTM & LR +Obj$^{\mathrm{c}}$ & 16 & 2 & 32 & Verbs & 32.81 & 73.701 & 12.84 & 10.41 & 898\\
        \hline
        7 & TSN \cite{wang_temporal_2016} & RGB stream & - & - & 25 & Verbs & 36.98 & 77.89 & 20.28 & 13.12 & 22 \\
        8 & TSN \cite{wang_temporal_2016} & Flow stream & - & - & 25 & Verbs & 37.99 & 76.45 & 23.14 & 14.06 & 22 \\
        9 & MF-Net \cite{chen_multi-fiber_2018} & RGB-3DConv & - & - & 16 & Verbs & \textbf{44.312} & \textbf{79.095} & 29.46 & 21.37 & 35 \\
        \hline
        \multicolumn{4}{l}{$^{\mathrm{a}}$Left/Right normalized hand coordinates (x,y)} & 
        \multicolumn{4}{l}{$^{\mathrm{b}}$Binary Presence Vector of detected objects} & 
        \multicolumn{4}{l}{$^{\mathrm{c}}$Normalized detected object coordinates (x,y)}\\
    \end{tabular}
    \label{tab:results}
\end{table*}

\subsection{LSTM Results}
\label{sec:lstm_results}
Initially, we test the ability of the LSTM to model the hand track sequences in full length using the LR feature. This is no straightforward task since the durations of action segments vary significantly from 0.5 seconds to 3.5 minutes which translates from 30 to as many as 12,000 frames. In our first experiment we train using the full hand coordinate sequences. For the LSTM to support training in batches with variable sequence sizes we zero-pad the shorter sequences to the size of the longest of the batch. For the shorter sequences we use the last hidden state before zero-padding as input to the FC layer and calculate the loss on this prediction. The Top1 accuracy is 31.1\% and the Top5 74.115\%. 

In an effort to simplify and speed-up the learning task, instead of smoothing the coordinates as in \cite{garg_drawinair:_2019}, we sample the action segments into shorter lengths. We are inspired from the concept used to train 3D CNNs, which aim to capture temporal structure, but due to computational restrictions are unable to load full frame sequences to represent video segments \cite{tran_learning_2015}.

\begin{figure}[htpb]
    \centering
    \subfloat[Original length]{\includegraphics[width=120pt]{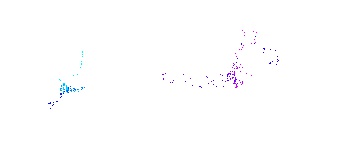}}
    \rule[0pt]{1pt}{50pt}
    \subfloat[Sampled to 32]{\includegraphics[width=119pt]{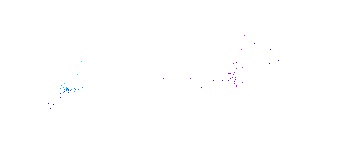}}
    \caption{Left and right hand motion patterns extracted from a 2.7s sequence for action `clean lid`. (a) The final view of the full sequence including all 161 steps. (b) The same sequence sampled to 32 steps. Zoom-in for best view.}
    \label{fig:fig3}
\end{figure}

In the second experiment, we sample the coordinate sequences to 32 steps and use these as input to LSTM. In Figure~\ref{fig:fig3} we visualize the difference between a full and a sampled sequence for the proposed sequence size. Furthermore, we reduce the number of hidden units per layer to avoid over-fitting, since the input is reduced significantly. Top1 performance drops \texttildelow1\% compared to the first experiment which can be attributed to the exclusion of temporal structure (more in Section~\ref{sec:discussion}).

For experiments three to five, we enhance the LR feature vector with object BPVs by appending them to the hand coordinates for every sequence step. For the "LR+BPV" experiments we incorporate the detected objects directly and for "LR+Trc BPV" we track the objects following the interpolation scheme of Section~\ref{sec:trc_interp}, in order to gain object presence knowledge for as many frames as possible. In experiment three we use the complete motion sequences (following experiment one) and improve Top1 classification accuracy by 3.8\% to 34.968\%. The addition of the BPV feature improves Top1 accuracy in the sampled sequences as well by 3\% to 34.053\% showing that the improvement from objects is consistent. Tracking the objects in the fifth experiment reaches 34.053\% without introducing further improvements.


In the sixth experiment, we use the tracked object coordinates to enhance the LR feature (LR+Obj) instead of the BPV. Again, we notice an improvement over having no object presence, however it is not as strong. We attribute it to the added uncertainty from location information about detections that may be false positives. The LSTM seems to be able to more adequately forget a falsely detected BPV from a coordinate that is propagated in the whole sequence.

\subsection{Comparison with video-based methods}
\label{sec:comparison_video}
In the final three experiments we compare against state-of-the-art video based methods, Temporal Segment Networks\cite{wang_temporal_2016} (TSN) and Multi-Fiber Networks\cite{chen_multi-fiber_2018} (MF-Net) that utilize Convolutional Neural Networks as feature extractors for two \cite{wang_temporal_2016} or three dimensional \cite{chen_multi-fiber_2018} inputs. In the 2D case, TSN extracts convolutional features from multiple stacks of either images (RGB stream) or pairs of horizontal and vertical optical flow values (Flow stream) that capture the perceived motion through series of images \cite{zach_duality_2007}. MF-Net utilizes a set of 16 frames sampled from the sequence of video frames to represent the segment. Both networks utilize only the video information without any contextual information about the scene. 

In terms of overall Top1 accuracy on the test set, the results are highest (44.3\%) when using 3D convolutions. Our methods remain close to TSN, but are still \texttildelow2\% lower. In Table~\ref{tab:perclass_analysis} we perform a class-wise comparison for the 10 most common verb classes in our train set, following the analysis in \cite{damen_scaling_2018}. We see that recall and precision are comparable between our methods and both TSN streams (experiments 1-4 and 7,8) for classes `put`, `take`, `wash`, `close`, `mix`, `pour` and `turn-on`. Against MF-Net we are close for actions `take`, `wash` and `pour`. This closeness in performance indicates an expressive quality in our data that can lead to action comprehension comparative to more elaborate methods by using only object detection and tracking as the means to deliver the input.
\begin{table*}[!ht]
    \caption{Comparison for the 10 most frequent verb classes in our training split. Showing  Recall and Precision per-class, results in \%}
    \centering
    \begin{tabular}{l|c@{\hspace{1pt}}:c@{\hspace{1pt}}|c@{\hspace{1pt}}:c@{\hspace{1pt}}|c@{\hspace{1pt}}:c@{\hspace{1pt}}|c@{\hspace{1pt}}:c@{\hspace{1pt}}|c@{\hspace{1pt}}:c@{\hspace{1pt}}|c@{\hspace{1pt}}:c@{\hspace{1pt}}|c@{\hspace{1pt}}:c@{\hspace{1pt}}|c@{\hspace{1pt}}:c@{\hspace{1pt}}|c@{\hspace{1pt}}:c@{\hspace{1pt}}|c@{\hspace{1pt}}:c@{\hspace{1pt}}}
         \# & \multicolumn{2}{c|}{put} & \multicolumn{2}{c|}{take} & \multicolumn{2}{c|}{wash} & \multicolumn{2}{c|}{open} & \multicolumn{2}{c|}{close} & \multicolumn{2}{c|}{cut} & \multicolumn{2}{c|}{mix} & \multicolumn{2}{c|}{pour} & \multicolumn{2}{c|}{move} & \multicolumn{2}{c}{turn-on} \\
         \cline{2-21}
         & R & P & R & P & R & P & R & P & R & P & R & P & R & P & R & P & R & P & R & P \\
         \hline
         1 & 42.36 & 33.29 & 48.55 & 28.16 & 63.09 & 36.78 & 11.85 & 26.5  & 12.9  & 19.13 & 24.35 & 45.16 & 13.07 & 30.3  & 33.82 & 18.85 & 0 & 0 & 0 & 0\\
         \hline
         2 & 64.53 & 28.89 & 40.57 & 29.1 & 52.89 & 39.6  & 2.87  & 55.56 & 0 & 0 & 14.78 & 47.89 & 27.45 & 25.61 & 0 & 0 & 0 & 0 & 0 & 0\\
         \hline
         3 & 42.05 & 34.78 & 56.61 & 31.57 & 68.32 & 43.62 & 18.93 & 39.92 & 16.72 & 27.4 & 28.26 & 34.03 & 43.14 & 38.6 &  8.82 & 13.64 & 0 & 0 & 2.2 & 50\\
         \hline
         4 & 34.53  & 34.48 & 68.58 & 28.66 & 62.55 & 44.42 & 17.02 & 37.87 & 12.9 & 36.67 & 20.87 & 47.06 & 28.1 & 55.13 & 1.47 & 4.55 & 0 & 0 & 1.1 & 12.5\\
         \hline
         7 & 57.24 & 29.89 & 43.19 & 29.8 & 58.93 & 60.3 & 43.21 & 51.36 & 12.61 & 39.45 & 54.35 & 67.57 & 30.72 & 51.09 & 10.29 & 26.92 & 0 & 0 & 2.2 & 20\\
         \hline
         8 & 31.87 & 38.95 & 81.81 & 29.9 & 37.05 & 56.33 & 58.51 & 58.4 & 24.34 & 62.88 & 23.91 & 60.44 & 41.18 & 70 & 32.35 & 28.57 & 0 & 0 & 0 & 0\\
         \hline
        9 & 52.62 & 39.55 & 60.88 & 37.47 & 62.01 & 63.2 & 55.64 & 53.49 & 26.39 & 54.88 & 60.43 & 64.35  & 53.59 & 61.19 & 26.47 & 36.73 & 0 & 0 & 24.18 & 31.43\\
    \end{tabular}
    \label{tab:perclass_analysis}
\end{table*}

\section{Discussion}
\label{sec:discussion}
\begin{figure}[bp]
    \centering
    \includegraphics[width=80pt]{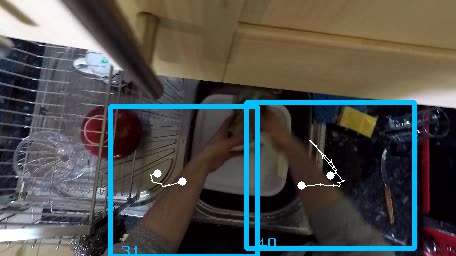}
    \includegraphics[width=80pt]{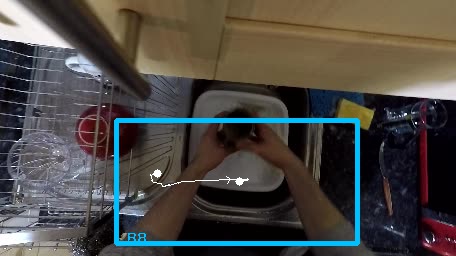}
    \includegraphics[width=80pt]{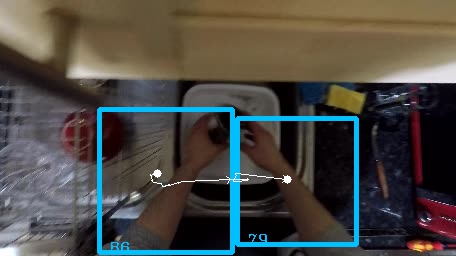}
    \caption{A double hand detection finds its way into the right hand track.}
    \label{fig:fig4}
\end{figure}

In this work we begin with a pipeline to capture the motions of regions of interest, in order to model the underlying human actions. This process is in close relation to the information it aims to comprehend and addresses specific issues that stem from detection and tracking in an egocentric setup. For example, in Figure~\ref{fig:fig0} we demonstrate a persistent complication with the hand detections that is successfully confronted with tracking. Another issue that may be introducing inconsistency to the hand tracks is the detection of both hands as one region of interest (\texttildelow2.5\% of detections). This leads to the assignment of the detection in either the left or the right hand track and momentarily produces an outlier coordinate (since the center of the detection is abruptly found elsewhere - see Figure~\ref{fig:fig4}). A way to suppress this could be to add a third "dual hand" track (e.g. LR+D as a feature) that captures these sequences and incorporates them in the final model as such; we will attempt this in our future work.

The contextual information added from the detection of objects, other than hands, contributes to the knowledge about actions, however we argue that improvement might be even higher with more accurate object detections. Previous research \cite{kapidis_where_2018, bambach_lending_2015} about the effect of using object annotations over detections supports this claim with a clear preference to minimizing misleading detections.

We view the process of standardizing all sequences to a certain length in experiments 2,4,5,6 as a manipulation of their temporal structure. After sampling, the \textit{temporal distance} between consecutive steps is not fixed to 16.7 ms (see also Section~\ref{sec:tracking}) but becomes a function of the sequence length and the sampling rate, which in turn originates from the amount of samples in the learning scheme and is not fixed for any two sequences. In essence we sacrifice part of the information related to the precise duration of each tiny motion step. The trade-off is significantly shorter training times per mini-batch (0.321s to 0.025s) and epoch (57.2s to 4.4s) in our setup with a 1080Ti GPU. In future work we plan to investigate spatial smoothing techniques (e.g. as in \cite{garg_drawinair:_2019}) instead of temporal to simplify the motions.

Hand tracks are our primary means for distinguishing human actions. Due to the high representative ability of human hands and their multipurpossness, the same motion can be expected to be part of multiple actions, for example `pull` and `take` are conceptually alike, hence the related hand motions are also expected to be similar. This includes an additional burden to our representation which we enhance using the objects in the scene. Investigating other sources of contextual information, such as the explicit duration of hand and object movements with an emphasis on hand-object interactions, together with improving existing sources through the removal of egomotion from the hand tracks and the enhancement of object detection are other directions we consider for future work.

\section{Conclusion}
\label{sec:conclusions}
In this work we perform a study regarding the usefulness of hand and object sequences for human action recognition. We focus on actions performed in kitchen environments, utilizing the recent Epic-Kitchens \cite{damen_scaling_2018} egocentric video dataset. We differentiate from state-of-the-art methods in activity recognition that utilize end-to-end video learning schemes with deep network structures in order to model explicitly the sequences of interest points detected in the scene. Our method comprises a detection and tracking scheme for the acquisition of hand motions from egocentric videos, which together with the detected objects in the scene are used to recognize egocentric actions as a sequence learning problem. 

Our results highlight the ability to infer a set of hand-based human actions with comparable accuracy to video-based methods, by only using a fraction of the input. In addition, we show that the inclusion of the presence of relevant detected objects enhances our feature set and improves performance.

This work is one of the few that comprehend that specialized hand movements can be interpreted as actions without the need to specifically rely on learned visual features for the temporal modeling. In our future work we plan to capitalize further on the human intelligible cues that define human actions.

\section*{Acknowledgment}
This project has received funding from the European Union's Horizon 2020 research and innovation program under the Marie Sk\l{}odowska-Curie grant agreement No 676157.

\bibliographystyle{ieeetr}
\bibliography{references1.bib}

\end{document}